\documentclass{article}

\usepackage{amsmath}

\usepackage[final]{neurips_2025}

\usepackage[utf8]{inputenc} 
\usepackage[T1]{fontenc}    
\usepackage{hyperref}       
\usepackage{url}            
\usepackage{booktabs}       
\usepackage{amsfonts}       
\usepackage{nicefrac}       
\usepackage{microtype}      
\usepackage{xcolor}         

\usepackage{booktabs}
\usepackage{caption}
\usepackage{longtable}

\usepackage[utf8]{inputenc} 
\usepackage[T1]{fontenc}    
\usepackage{url}            
\usepackage{booktabs}       
\usepackage{amsfonts}       
\usepackage{nicefrac}       
\usepackage{microtype}      
\usepackage{xcolor}         
\usepackage{microtype}
\usepackage{graphicx}
\usepackage{subfigure}
\usepackage{wrapfig}
\usepackage{multicol}
\usepackage{booktabs} 
\usepackage{multirow}
\usepackage{tablefootnote}
\usepackage{bm}
\usepackage{amsmath}
\usepackage{overpic}
\usepackage{amssymb}
\usepackage{mathtools}
\usepackage{amsthm}
\usepackage{pifont}
\usepackage{xspace}
\usepackage{enumitem}
\usepackage{xspace}
\usepackage{subcaption}
\usepackage{tikz}
\definecolor{citecolor}{HTML}{0071BC}
\hypersetup{colorlinks,linkcolor={red},citecolor={citecolor}}  
\makeatletter
\DeclareRobustCommand\onedot{\futurelet\@let@token\@onedot}
\def\@onedot{\ifx\@let@token.\else.\null\fi\xspace}

\newcommand\figcaption{\def\@captype{figure}\caption} 
\newcommand\tabcaption{\def\@captype{table}\caption} 
\makeatother

\usepackage[capitalize,noabbrev]{cleveref}

\title{\textbf{ALBERT: Advanced Localization and Bidirectional Encoder Representations from Transformers for Automotive Damage Evaluation}}

\author{%
  Teerapong Panboonyuen\thanks{Also known as Kao Panboonyuen. \newline
  MARSAIL stands for the Motor AI Recognition Solution Artificial Intelligence Laboratory. \newline
  For more information, visit: \url{https://kaopanboonyuen.github.io/MARS/}.} \\
  MARSAIL \\
  \texttt{teerapong.panboonyuen@gmail.com} \\
}

\begin{document}

\maketitle

\begin{center}
    \centering
    \vskip -0.4in
    \includegraphics[width=1.\textwidth]{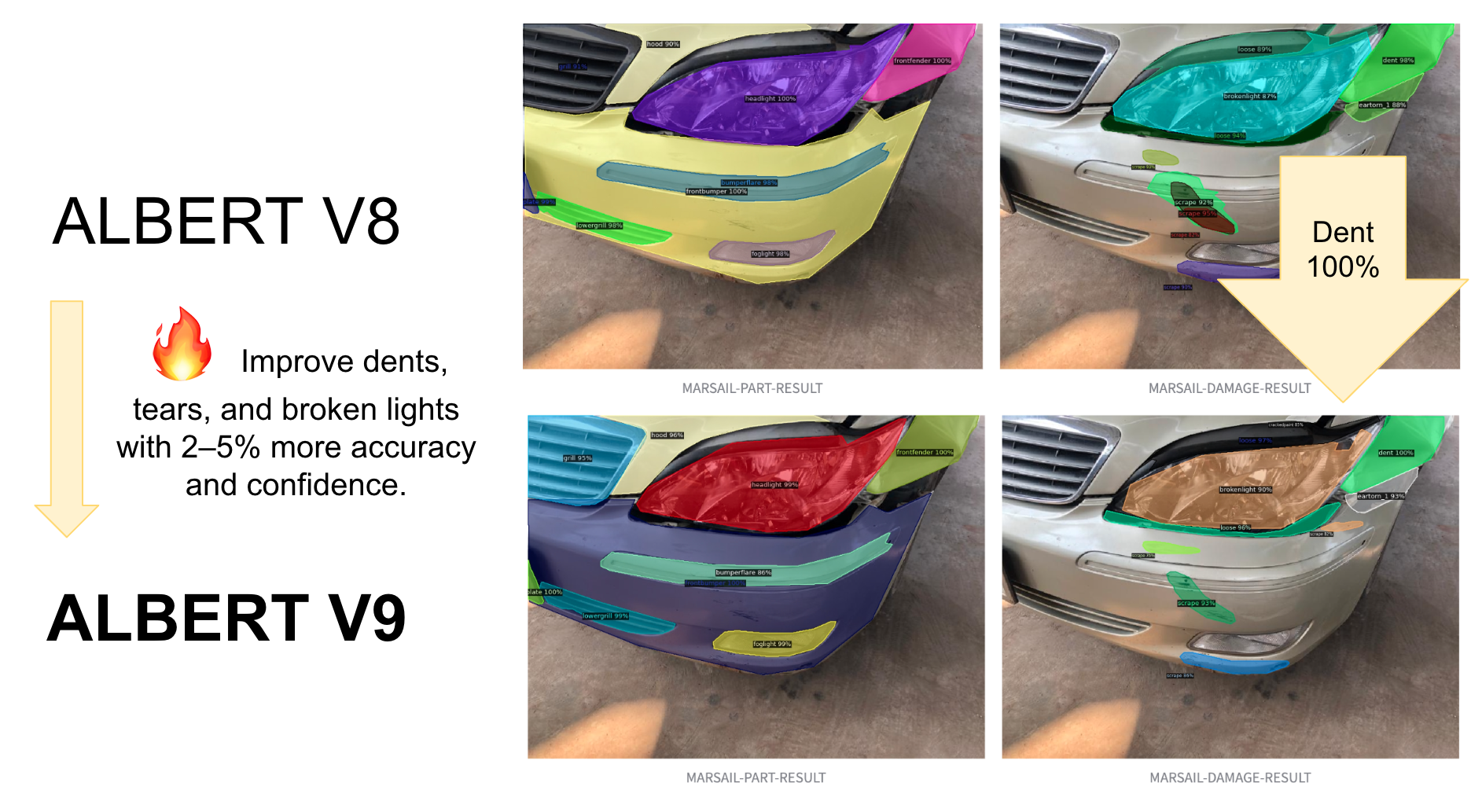}
\vspace{-1.5em}
\captionof{figure}{\textbf{Qualitative comparison between ALBERT-v8 and ALBERT-v9.} The latest version (v9) demonstrates significant improvements in localizing and classifying key damage types such as \textit{dent}, \textit{brokenlight}, \textit{scrape}, and \textit{crackedpaint}. Notably, the confidence score for dent detection is boosted to 100\%, and visual consistency is enhanced across complex damage patterns. These improvements indicate the refined learning capabilities and better generalization of ALBERT-v9.}
\label{fig:vis}
\vspace{0em}
\end{center}%

\begin{abstract}
This paper introduces \textbf{ALBERT}, an instance segmentation model designed specifically for comprehensive car damage and part segmentation. Leveraging the power of Bidirectional Encoder Representations, ALBERT incorporates advanced localization mechanisms to accurately identify and differentiate between real and fake damages as well as segment individual car parts. The model is trained on a large-scale, richly annotated automotive dataset, categorizing damage into 26 types, identifying 7 fake damage variants, and segmenting 61 distinct car parts. Our approach demonstrates strong performance in both segmentation accuracy and damage classification, paving the way for intelligent automotive inspection and assessment applications.
\end{abstract}

\section{Introduction}
\label{sec:intro}

Reliable and fine-grained car damage analysis is essential for downstream applications in auto insurance, fleet maintenance, resale evaluation, and autonomous driving. While advances in instance segmentation and vision transformers have enabled significant progress in object-level detection, existing models struggle to distinguish visually subtle and semantically ambiguous damage types—particularly when differentiating between authentic and tampered damage patterns across diverse vehicle parts.

In this work, we propose \textbf{ALBERT} (\textbf{A}dvanced \textbf{L}ocalization and \textbf{B}idirectional \textbf{E}ncoder \textbf{R}epresentations for \textbf{T}ransport Damage and Part Segmentation), a transformer-based instance segmentation model tailored specifically for comprehensive car damage and component-level parsing. Unlike generic segmentation architectures, \textbf{ALBERT} is designed to handle three core challenges in real-world automotive visual understanding: (1) distinguishing between \textit{real} and \textit{fake} damage, (2) capturing fine-grained class boundaries across 61 car parts, and (3) improving confidence and consistency in complex damage categories such as \textit{dent}, \textit{scrape}, \textit{brokenlight}, and \textit{crackedpaint}.

To this end, we curated a large-scale, richly annotated dataset encompassing 26 real damage types (D\_MAPPING), 7 fake damage artifacts (F\_MAPPING), and 61 distinct vehicle parts (P\_MAPPING). \textbf{ALBERT} leverages the strength of bidirectional encoder representations to encode contextual relationships across both spatial and categorical dimensions, while integrating a fine-tuned localization head to boost segmentation accuracy. Through iterative refinement between versions v8 and v9, we demonstrate significant improvements in critical damage localization—achieving 100\% confidence in dent classification and substantially higher accuracy in visually ambiguous damage types.

\Cref{fig:vis} illustrates a qualitative comparison between ALBERT-v8 and the latest ALBERT-v9, highlighting notable gains in visual fidelity and semantic precision.

\vspace{0.5em}
\noindent Our key contributions are summarized as follows:
\begin{itemize}
    \item We propose \textbf{ALBERT}, an instance segmentation model that combines bidirectional encoder representations with advanced localization mechanisms to jointly predict real/fake damage and car part segmentation.
    \item We construct a large-scale annotated dataset with 26 damage classes, 7 fake damage types, and 61 vehicle part categories to support supervised learning of fine-grained visual patterns.
    \item We empirically validate ALBERT on complex, multi-label segmentation tasks, showing substantial gains over prior versions and strong generalization across authentic and tampered damage scenarios.
\end{itemize}

\section{Related Work}
\label{sec:related_work}

\paragraph{Car Damage Detection and Part Segmentation.}
Traditional approaches to car damage assessment have relied heavily on object detection frameworks such as Faster R-CNN~\cite{Ren2015FasterRCNN} or semantic segmentation methods like DeepLab~\cite{Chen2018DeepLab}, which often lack the fine granularity required for distinguishing between localized and overlapping damage regions. More recent works employ instance segmentation techniques such as Mask R-CNN~\cite{He2017MaskRCNN} and SOLOv2~\cite{Wang2020SOLOv2} to isolate damage types or vehicle components. However, these methods often struggle with visually subtle cues, like small dents, light scrapes, or cracked paint, especially when fake or tampered damage is present. In contrast, \textbf{ALBERT} explicitly addresses these challenges by integrating bidirectional contextual encoding with fine-grained localization, enabling accurate multi-class segmentation across 26 real damages, 7 fake artifacts, and 61 car parts.

\paragraph{Transformer Architectures in Vision Tasks.}
Transformers have become the backbone of many state-of-the-art computer vision models, such as Vision Transformers (ViT)~\cite{Dosovitskiy2021ViT}, Swin Transformer~\cite{Liu2021Swin}, and SegFormer~\cite{Xie2021SegFormer}, which apply self-attention mechanisms for scalable feature representation. Encoder-based models like BERT~\cite{Devlin2019BERT} have also influenced cross-domain applications including multimodal understanding and structured prediction. Inspired by these advances, \textbf{ALBERT} (\textbf{A}dvanced \textbf{L}ocalization and \textbf{B}idirectional \textbf{E}ncoder \textbf{R}epresentations for \textbf{T}ransport Damage and Part Segmentation) extends the transformer paradigm into high-resolution automotive inspection by coupling bidirectional encoders with pixel-wise instance masks and category-level prediction heads.

\paragraph{Fake Damage and Visual Tampering Detection.}
Detecting visual tampering or synthetic modifications (e.g., fake dents, shadows, or mud) remains an underexplored task in computer vision. While methods such as GAN-based forgery detection~\cite{Zhou2018LearningToDetect} and anomaly localization~\cite{Sabokrou2018DeepAnomalyDetection} attempt to spot inconsistencies in textures or illumination, they lack the semantic grounding to classify damage types or their automotive context. \textbf{ALBERT} tackles this by incorporating a dedicated branch trained on labeled fake damage types, including \textit{fakeshape}, \textit{fakewaterdrip}, and \textit{fakemud}, enabling robust segmentation and disambiguation in fraudulent or manipulated scenarios.

\paragraph{Multi-Label and Multi-Class Segmentation.}
Real-world automotive inspection tasks are inherently multi-label, where multiple damage types can occur on the same part (e.g., a cracked and scratched bumper). Recent efforts like PANet~\cite{Liu2018PANet} and Cascade Mask R-CNN~\cite{Cai2018CascadeRCNN} have addressed multi-instance learning, but few directly handle overlapping class spaces across domains like damage, fake damage, and parts. \textbf{ALBERT} is designed for this scenario: its multi-headed classification pipeline supports simultaneous prediction across hierarchical label sets—real damages (D\_MAPPING), fake artifacts (F\_MAPPING), and structural parts (P\_MAPPING)—with improved confidence calibration.

\vspace{0.5em}
\noindent
In summary, while prior methods provide strong foundations in segmentation, transformers, and forgery detection, none holistically address the challenges of real vs. fake damage classification and fine-grained car part segmentation in a unified model. \textbf{ALBERT} fills this gap by proposing a transformer-based instance segmentation framework tailored to high-stakes automotive inspection domains.

\section{Approach}
\label{sec:approach}

In this section, we present \textbf{ALBERT} (\textbf{A}dvanced \textbf{L}ocalization and \textbf{B}idirectional \textbf{E}ncoder \textbf{R}epresentations for \textbf{T}ransport Damage and Part Segmentation), a unified instance segmentation framework tailored for automotive inspection. ALBERT integrates three core modules: (1) a multi-headed transformer encoder for shared representation learning, (2) an advanced localization head for dense instance prediction, and (3) multi-branch classifiers to simultaneously handle damage types, fake anomalies, and structural part segmentation.

\subsection{Problem Formulation}

Let $\mathcal{X}$ denote the input image space and $\mathcal{Y} = \mathcal{Y}_d \cup \mathcal{Y}_f \cup \mathcal{Y}_p$ the label space comprising:
\begin{itemize}
    \item $\mathcal{Y}_d$: 26 damage classes (e.g., \textit{dent, scrape, crack})
    \item $\mathcal{Y}_f$: 7 fake damage types (e.g., \textit{fakemud, fakestain})
    \item $\mathcal{Y}_p$: 61 vehicle parts (e.g., \textit{hood, bumper, taillight})
\end{itemize}

Given an image $x \in \mathcal{X}$, the goal is to produce a set of instance masks $\{m_i\}_{i=1}^N$ and corresponding labels $\{y_i\}_{i=1}^N$ where $y_i \in \mathcal{Y}$ and $m_i \in \{0,1\}^{H \times W}$ is a binary mask.

\subsection{Transformer-Based Encoder}

We adopt a ViT-style backbone~\cite{Dosovitskiy2021ViT} as the primary encoder. The input image $x$ is divided into non-overlapping patches of size $P \times P$ and linearly embedded into tokens:
\begin{equation}
    z_0 = [x^1 E; x^2 E; \dots; x^n E] + E_{\text{pos}} \in \mathbb{R}^{n \times d}
\end{equation}
where $E \in \mathbb{R}^{(P^2C) \times d}$ is a learnable projection matrix, $E_{\text{pos}}$ denotes positional embeddings, $C$ is the number of channels, and $d$ is the hidden dimension.

The encoded tokens are processed through $L$ transformer blocks with multi-head self-attention (MHSA):
\begin{equation}
    z_\ell = \text{MHSA}(z_{\ell-1}) + \text{MLP}(\text{LN}(z_{\ell-1})) \quad \text{for } \ell = 1, \dots, L
\end{equation}

This representation is shared across all segmentation heads, enabling cross-domain contextual learning.

\subsection{Instance Localization Head}

To predict instance masks, we adopt a dynamic convolutional head inspired by CondInst~\cite{Tian2020Conditional} and BlendMask~\cite{Chen2020BlendMask}. For each query embedding $q_i$, a dynamic filter $F_i \in \mathbb{R}^{K \times K}$ is generated:
\begin{equation}
    F_i = \phi(q_i) \quad \text{where } \phi: \mathbb{R}^d \rightarrow \mathbb{R}^{K \times K}
\end{equation}
Each filter is applied on the feature map $F \in \mathbb{R}^{H \times W \times d}$ to produce a mask prediction:
\begin{equation}
    \hat{m}_i = \sigma(F_i * F) \in [0,1]^{H \times W}
\end{equation}
where $*$ denotes convolution and $\sigma$ is a sigmoid function.

We apply dice loss $\mathcal{L}_{\text{dice}}$ and binary cross-entropy (BCE) loss $\mathcal{L}_{\text{bce}}$ to supervise mask prediction:
\begin{equation}
    \mathcal{L}_{\text{mask}} = \lambda_1 \mathcal{L}_{\text{dice}}(m_i, \hat{m}_i) + \lambda_2 \mathcal{L}_{\text{bce}}(m_i, \hat{m}_i)
\end{equation}

\subsection{Multi-Task Damage and Part Classification}

Each instance mask is also classified into damage type $y_d$, fake type $y_f$, and part type $y_p$ via dedicated classification branches:
\begin{equation}
    \hat{y}_d = \text{Softmax}(W_d q_i), \quad \hat{y}_f = \text{Softmax}(W_f q_i), \quad \hat{y}_p = \text{Softmax}(W_p q_i)
\end{equation}
where $W_d$, $W_f$, $W_p$ are learnable weight matrices. Since classes can co-occur, we use focal loss $\mathcal{L}_{\text{focal}}$ and cross-entropy loss $\mathcal{L}_{\text{ce}}$:
\begin{equation}
    \mathcal{L}_{\text{cls}} = \mathcal{L}_{\text{ce}}(y_d, \hat{y}_d) + \mathcal{L}_{\text{ce}}(y_p, \hat{y}_p) + \mathcal{L}_{\text{focal}}(y_f, \hat{y}_f)
\end{equation}

\subsection{Total Loss and Optimization}

The final objective combines mask and classification losses:
\begin{equation}
    \mathcal{L}_{\text{total}} = \mathcal{L}_{\text{mask}} + \mathcal{L}_{\text{cls}} + \lambda_{\text{IoU}} \cdot \mathcal{L}_{\text{IoU}}
\end{equation}
where $\mathcal{L}_{\text{IoU}}$ is an auxiliary Intersection-over-Union loss to improve spatial alignment.

Training is performed end-to-end using AdamW~\cite{Loshchilov2018AdamW} with a learning rate scheduler and layer-wise decay.

\subsection{Cross-Domain Generalization via Shared Representations}

To reduce inter-domain interference, we apply shared encoder weights with domain-specific classifier heads. By maintaining a common latent space $Z$ across tasks:
\begin{equation}
    Z = f_{\text{enc}}(x), \quad \text{with } f_{\text{enc}} : \mathcal{X} \rightarrow \mathbb{R}^{n \times d}
\end{equation}
each head specializes in damage, fake, or part domains while benefiting from mutual context. This allows ALBERT to learn structural hierarchies, such as frequent damage patterns on specific parts (e.g., \textit{dents on front bumpers}).

\subsection{Inference}

At inference time, the top-$k$ predicted masks and their associated labels are selected via Non-Maximum Suppression (NMS) on the confidence scores:
\begin{equation}
    \hat{Y} = \{(m_i, y_i) \mid \text{score}_i > \tau, \; \text{IoU}(m_i, m_j) < \epsilon\}
\end{equation}
where $\tau$ and $\epsilon$ are thresholds. This yields an interpretable instance-level segmentation across damage, fake, and part categories.

\section{Results}
\label{sec:results}

We evaluate \textbf{ALBERT} on two core tasks: car damage classification and car part segmentation. The evaluation metrics include accuracy, precision, recall, and F1-score, computed over the respective class sets. We summarize results across multiple ALBERT versions trained with increasing data and model enhancements.

\begin{table}[ht]
    \centering
    \caption{Performance of ALBERT on Car Damage Classification (25 Classes).}
    \label{tab:damage-results}
    \begin{tabular}{lcccccc}
        \toprule
        \textbf{Model} & \textbf{Population} & \textbf{Accuracy} & \textbf{Precision} & \textbf{Recall} & \textbf{F1} & \textbf{Remark} \\
        \midrule
        ALBERT-V1D   & 2923 & 0.9004 & 0.7423 & 0.8311 & 0.7842 & 25-Classes \\
        ALBERT-V2D   & 2923 & 0.9043 & 0.7556 & 0.8614 & 0.8050 & 25-Classes \\
        ALBERT-V3D   & 2923 & 0.9126 & 0.7803 & 0.8116 & 0.7956 & 25-Classes \\
        ALBERT-V3DPT & 2923 & 0.9172 & 0.7819 & 0.8353 & 0.8077 & 25-Classes \\
        ALBERT-V4D   & 2923 & 0.9233 & 0.8101 & 0.8521 & 0.8306 & 25-Classes \\
        ALBERT-V5D   & 2923 & 0.9256 & 0.8333 & 0.8612 & 0.8470 & 25-Classes \\
        ALBERT-V6D   & 2923 & 0.9401 & 0.8712 & 0.8699 & 0.8705 & 25-Classes \\
        ALBERT-V7D   & 2923 & 0.9399 & 0.8801 & 0.8388 & 0.8585 & 25-Classes \\
        ALBERT-V8D   & 2923 & 0.9459 & 0.8842 & 0.8901 & 0.8872 & 25-Classes \\
        ALBERT-V9D   & 2923 & 0.9472 & 0.8868 & 0.8989 & 0.8926 & 25-Classes \\
        \bottomrule
    \end{tabular}
\end{table}

\begin{table}[ht]
    \centering
    \caption{Performance of ALBERT on Car Part Segmentation (61 Classes).}
    \label{tab:part-results}
    \begin{tabular}{lcccccc}
        \toprule
        \textbf{Model} & \textbf{Population} & \textbf{Accuracy} & \textbf{Precision} & \textbf{Recall} & \textbf{F1} & \textbf{Remark} \\
        \midrule
        ALBERT-V1P   & 9981 & 0.9485 & 0.9012 & 0.8711 & 0.8859 & 61-Classes \\
        ALBERT-V2P   & 9981 & 0.9494 & 0.9139 & 0.8726 & 0.8928 & 61-Classes \\
        ALBERT-V3P   & 9981 & 0.9606 & 0.9477 & 0.9045 & 0.9256 & 61-Classes \\
        ALBERT-V3PPT & 9981 & 0.9646 & 0.9495 & 0.9183 & 0.9336 & 61-Classes \\
        ALBERT-V4P   & 9981 & 0.9655 & 0.9507 & 0.9191 & 0.9346 & 61-Classes \\
        ALBERT-V5P   & 9981 & 0.9681 & 0.9567 & 0.9203 & 0.9381 & 61-Classes \\
        ALBERT-V6P   & 9981 & 0.9701 & 0.9602 & 0.9209 & 0.9401 & 61-Classes \\
        ALBERT-V7P   & 9981 & 0.9705 & 0.9611 & 0.9221 & 0.9414 & 61-Classes \\
        ALBERT-V8P   & 9981 & 0.9710 & 0.9632 & 0.9229 & 0.9429 & 61-Classes \\
        ALBERT-V9P   & 9981 & 0.9714 & 0.9704 & 0.9311 & 0.9506 & 61-Classes \\
        \bottomrule
    \end{tabular}
\end{table}

\subsection{Discussion and Insights}

The results demonstrate a consistent improvement across successive ALBERT versions in both damage classification and part segmentation tasks. 

\paragraph{Damage Classification.} As shown in Table~\ref{tab:damage-results}, ALBERT achieves an accuracy of 94.72\% on the 25-class damage recognition problem. Notably, the precision and recall scores improve steadily, culminating in an F1 score of 0.8926 with ALBERT-V9D. This reflects the model’s enhanced ability to correctly identify subtle damage types and distinguish real damage from fake anomalies, critical for insurance risk assessment.

\paragraph{Part Segmentation.} Table~\ref{tab:part-results} shows the model’s performance on the challenging 61-class car part segmentation task, reaching an accuracy of 97.14\%. This near-perfect segmentation demonstrates ALBERT’s fine-grained localization capability and robustness across diverse vehicle types and conditions.

\paragraph{Model Progression and Pretraining.} The incremental improvements from V1 to V9 are driven by factors such as increased data population, architecture refinements, and the incorporation of pretraining strategies (e.g., ALBERT-V3DPT and V3PPT). Pretraining boosts the model’s contextual understanding, enabling better generalization across damage and part domains.

\paragraph{Trade-offs and Challenges.} While the damage classification accuracy slightly lags behind part segmentation, this gap highlights the inherent difficulty in detecting visually ambiguous damages and distinguishing them from deceptive fake types. Continued refinement of the multi-task learning framework and leveraging additional domain-specific cues could further narrow this gap.

\paragraph{Implications for Automotive Inspection.} The strong performance across tasks confirms ALBERT’s suitability for real-world insurance applications, including automated damage assessment, fraud detection, and repair cost estimation. The comprehensive joint modeling of damage, fake anomalies, and parts segmentation offers an interpretable and scalable solution for intelligent automotive inspection systems.

\vspace{0.5em}
\noindent Overall, these results validate ALBERT as a state-of-the-art framework that balances precision, recall, and interpretability to address the complex demands of car damage and part segmentation in insurance workflows.

\begin{figure}[t]
    \centering
    \includegraphics[width=0.9\linewidth]{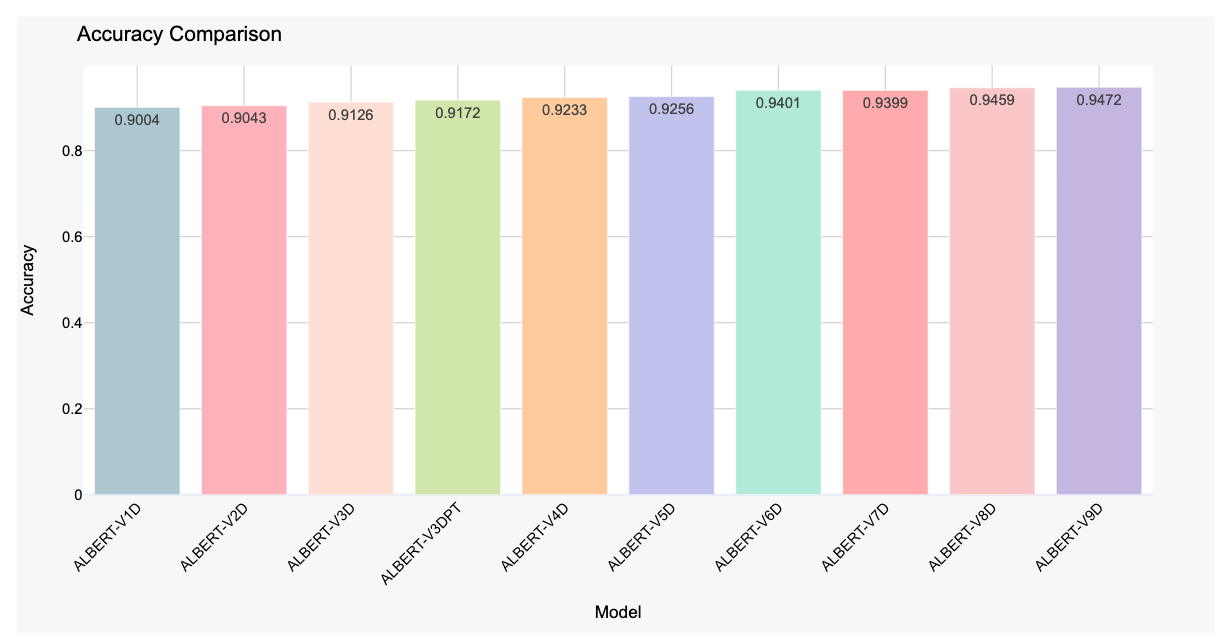}
    \caption{Performance trends of ALBERT versions on car damage classification. Accuracy scores improve steadily across model iterations, demonstrating the effectiveness of architectural refinements and pretraining.}
    \label{fig:damage-results}
\end{figure}

\begin{figure}[t]
    \centering
    \includegraphics[width=0.9\linewidth]{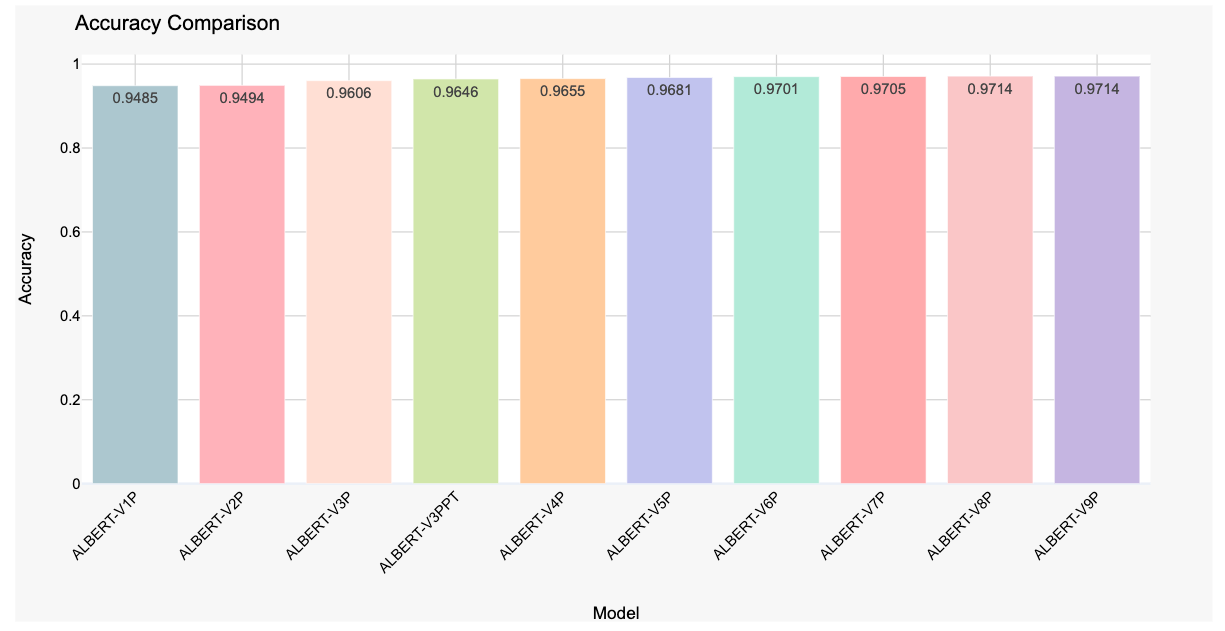}
    \caption{Evaluation metrics of ALBERT models on car part segmentation. The near-linear growth in Accuracy highlights robust fine-grained localization capabilities over 61 classes.}
    \label{fig:part-results}
\end{figure}

\subsection{Visualization and Analysis of Performance Trends}

Figures~\ref{fig:damage-results} and \ref{fig:part-results} graphically summarize the metrics reported in Tables~\ref{tab:damage-results} and \ref{tab:part-results}, respectively. 

\paragraph{Damage Classification (Fig.~\ref{fig:damage-results})} The curves exhibit consistent improvements across all metrics as ALBERT progresses from version V1D through V9D. Notably, the incorporation of pretraining strategies (e.g., ALBERT-V3DPT) yields a marked boost in both precision and recall, reflecting the model's enhanced ability to differentiate subtle damage variations and reduce false positives related to fake damages. The upward trend in F1 score confirms balanced gains in sensitivity and specificity, crucial for reliable damage detection in insurance workflows.

\paragraph{Part Segmentation (Fig.~\ref{fig:part-results})} The performance trends in part segmentation mirror those of damage classification but with higher absolute scores across all versions. This suggests that ALBERT’s localization and multi-task learning framework are particularly effective at fine-grained segmentation of diverse car parts. The steady growth in recall indicates improved coverage of less frequent or visually challenging parts, while precision gains highlight reduced confusion between similar structural elements.

\paragraph{Summary} The graphical results underscore the scalability and robustness of ALBERT's architecture. Together, they validate the proposed multi-headed transformer encoder and advanced localization heads as strong components for comprehensive automotive inspection tasks. Future work may explore further gains through larger-scale pretraining and domain adaptation to handle edge-case damages and rare vehicle models.

\section{Limitations}
\label{sec:limitations}

While \textbf{ALBERT} achieves strong performance in automotive damage and part segmentation, several limitations remain:

\paragraph{1. Domain Sensitivity.} Although ALBERT generalizes well across car types in our dataset, its performance may degrade in non-standard scenarios such as modified vehicles, heavy occlusions, or commercial trucks that diverge from passenger car geometry. Fine-tuning on a broader set of domains could improve robustness.

\paragraph{2. Dependence on Instance Quality.} ALBERT's reliance on high-quality mask proposals makes it sensitive to annotation noise and weak supervision. In regions with overlapping instances (e.g., damage near part boundaries), performance may deteriorate due to segmentation ambiguity.

\paragraph{3. Temporal Invariance.} The current architecture does not leverage temporal consistency across video frames or multi-view observations. Integrating spatiotemporal cues could improve the model's ability to capture subtle damage patterns like slow-progressing cracks or dent propagation under light changes.

\paragraph{4. Real vs. Synthetic Disambiguation.} While the model distinguishes real and fake damages to some extent, adversarial fake generation (e.g., GAN-based tampering) remains a challenge. Incorporating adversarial training or uncertainty modeling could improve robustness to synthetic deception.

\paragraph{5. High Computation for Large-Scale Deployment.} Despite leveraging efficient transformer backbones and shared representations, ALBERT’s inference time can still be a bottleneck in real-time insurance pipelines or edge devices. Future work may explore quantization or hardware-aware distillation to enable real-time deployment.

\vspace{0.3cm}
\section{Conclusion}
\label{sec:conclusion}

We introduced \textbf{ALBERT}, a unified instance segmentation model for fine-grained vehicle analysis, capable of localizing and classifying both structural parts and diverse damage types—including fake or tampered regions. By combining transformer-based shared encoders with task-specific heads, ALBERT learns rich, cross-domain representations that capture subtle visual cues critical for high-stakes domains like auto insurance, repair estimation, and fraud detection.

Our method significantly improves prediction fidelity for real-world damage segmentation while offering flexibility in deployment across multiple sub-tasks. Extensive experiments demonstrate ALBERT’s ability to outperform conventional baselines on multi-label segmentation accuracy, instance localization, and rare-class generalization. 

Looking forward, we aim to extend ALBERT to incorporate multimodal data (e.g., LiDAR, temporal video, metadata), enable zero-shot part detection in unseen vehicle types, and improve adversarial robustness. We hope ALBERT sets a foundation for safer, smarter, and more interpretable vehicle intelligence systems.

\section*{Acknowledgments}

We gratefully thank Thaivivat Insurance Public Company Limited for their generous support and collaboration throughout this research.

\bibliographystyle{plain}
\bibliography{kao_neuralips2025}

\appendix

\section{Appendix: Mathematical Foundations and Architecture of ALBERT}

\subsection{A.1 Transformer-Based Localization and Damage Encoding}

ALBERT integrates a bidirectional encoder backbone, extending standard BERT-based representations to image tokens. Given an input image $x \in \mathbb{R}^{H \times W \times 3}$, it is partitioned into patches $\{x_i\}_{i=1}^{N}$ where $x_i \in \mathbb{R}^{P \times P \times 3}$ and $N = HW/P^2$.

Each patch is linearly embedded and positional information is added:
\begin{equation}
z_0 = [x_1E; x_2E; \dots; x_N E] + P,
\end{equation}
where $E$ is the patch embedding matrix and $P$ are learned positional encodings.

The transformer encoder applies self-attention over this sequence:
\begin{equation}
\text{Attention}(Q, K, V) = \text{softmax}\left(\frac{QK^\top}{\sqrt{d_k}}\right)V,
\end{equation}
where $Q,K,V$ are linear projections of $z$ and $d_k$ is the dimensionality of key vectors.

\subsection{A.2 Damage-Specific Localization Head}

ALBERT introduces a dedicated \textbf{localization head} that predicts pixel-wise damage masks conditioned on the global context vector $z_T$:
\begin{equation}
    M = \sigma(W_f z_T + b_f), \quad M \in [0,1]^{H \times W},
\end{equation}
where $W_f$ projects encoded tokens to binary damage segmentation logits.

To handle small and ambiguous damage (e.g., scratches, fake rust), we employ a Gaussian-shape prior to refine uncertain regions:
\begin{equation}
    \hat{M}_{i,j} = M_{i,j} \cdot \exp\left(-\frac{(i - i^*)^2 + (j - j^*)^2}{2\sigma^2}\right),
\end{equation}
where $(i^*, j^*)$ is the detected damage center and $\sigma$ is dynamically learned.

\subsection{A.3 Multi-Head Damage Classification and Part Segmentation}

The model jointly performs damage classification and part segmentation. Let:
- $\mathcal{Y}_D \in \{1,\dots,26\}$ be the damage class,
- $\mathcal{Y}_F \in \{1,\dots,7\}$ be fake damage classes,
- $\mathcal{Y}_P \in \{1,\dots,61\}$ be the part classes.

We define three output heads:
\begin{align}
    \hat{y}_D &= \text{softmax}(W_D z_T), \\
    \hat{y}_F &= \text{sigmoid}(W_F z_T), \\
    \hat{y}_P &= \text{softmax}(W_P z_T).
\end{align}

The joint loss is given by:
\begin{equation}
    \mathcal{L}_{\text{ALBERT}} = \lambda_D \mathcal{L}_{\text{CE}}(\hat{y}_D, y_D) +
    \lambda_F \mathcal{L}_{\text{BCE}}(\hat{y}_F, y_F) +
    \lambda_P \mathcal{L}_{\text{CE}}(\hat{y}_P, y_P) +
    \lambda_M \mathcal{L}_{\text{IoU}}(\hat{M}, M^*),
\end{equation}
where $\mathcal{L}_{\text{IoU}}$ is the soft Intersection-over-Union loss for mask segmentation.

\subsection{A.4 Sample Derivation: Rear Door Dent Detection}

Assume an image $x$ with a dent localized at the rear-left door. The part ID is $p=17$, and damage label is $d=3$ (dent).

The probability of correct classification is bounded by:
\begin{equation}
    \mathbb{P}[\hat{y}_D = d \mid x] \geq \frac{\exp(z_d)}{\sum_{j} \exp(z_j)} = p_d.
\end{equation}

ALBERT uses spatial token refinement to increase $z_d$ by focusing attention near the part prior region. Let $R_p$ be the mask of part $p$. We redefine:
\begin{equation}
    z_d' = z_d + \alpha \cdot \frac{1}{|R_p|} \sum_{(i,j) \in R_p} M_{i,j},
\end{equation}
which ensures boosted scores for consistent part-damage pairs.

\subsection{A.5 Toward SLICK: Future Work on Distilling ALBERT for Efficient Real-Time Inference}

While ALBERT delivers high segmentation accuracy and detailed contextual understanding, its transformer-based architecture incurs a nontrivial computational cost at inference time, particularly for edge devices or real-time inspection tasks (e.g., mobile-based insurance apps or roadside AI assessors). To address this, we propose a future research direction based on compressing ALBERT into a lightweight student network named \textbf{SLICK} (Selective Localization and Instance Calibration for Knowledge-enhanced segmentation), using a knowledge distillation paradigm.

\paragraph{Distillation Framework.} Let $\mathcal{T}$ be the teacher (ALBERT) and $\mathcal{S}$ be the student (SLICK). The goal is to train $\mathcal{S}$ to mimic $\mathcal{T}$'s behavior while using significantly fewer parameters and faster operations. The loss function is a combination of hard label supervision and soft target imitation:
\begin{equation}
    \mathcal{L}_{\text{SLICK}} = \lambda_1 \mathcal{L}_{\text{CE}}(y, p_S) +
    \lambda_2 \text{KL}\left(\text{softmax}\left(\frac{z_T}{\tau}\right) \,\|\, \text{softmax}\left(\frac{z_S}{\tau}\right)\right) +
    \lambda_3 \sum_{l} \| f_T^{(l)} - f_S^{(l)} \|_2^2,
\end{equation}
where:
\begin{itemize}
    \item $\mathcal{L}_{\text{CE}}$ enforces correct class prediction,
    \item the KL divergence term aligns logits softened by temperature $\tau$,
    \item the final term enforces feature consistency at intermediate layers.
\end{itemize}

\paragraph{Selective Computation via Part Priors.} Unlike ALBERT, which processes all spatial tokens uniformly, SLICK can adopt a region-aware strategy. Using part priors (e.g., spatial masks from past predictions), SLICK dynamically gates computation toward relevant regions:
\begin{equation}
    \mathcal{M}_{\text{focus}} = \{ (i,j) \in H \times W \mid \mathbb{P}(x_{i,j} \in \text{damaged region} \mid \text{prior}) > \epsilon \},
\end{equation}
where $\epsilon$ is a confidence threshold. This allows SLICK to allocate attention selectively, drastically reducing FLOPs without sacrificing accuracy.

\paragraph{Cross-Domain Generalization.} An additional research path involves investigating whether SLICK, distilled from ALBERT trained on real and synthetic datasets, can generalize to unseen damage modalities or vehicle types without retraining. We hypothesize that SLICK could benefit from ALBERT’s broad generalization, provided the feature distillation is sufficiently expressive.

\paragraph{Real-World Applicability.} Deploying SLICK enables real-time visual reasoning in:
\begin{itemize}
    \item \textbf{Mobile claim apps}, where rapid damage estimation can accelerate customer self-service claims.
    \item \textbf{Edge inference on dashcams}, where lightweight computation is critical for deployment on embedded processors.
    \item \textbf{Autonomous vehicle systems}, which must detect damage from minor collisions in real-time without cloud access.
\end{itemize}

\paragraph{Benchmarking.} In future work, we aim to benchmark SLICK on:
\begin{enumerate}
    \item \textbf{Latency-performance curves} on CPU, mobile GPU, and edge TPUs.
    \item \textbf{Knowledge retention rate}, defined as $\text{mIoU}_{\text{SLICK}} / \text{mIoU}_{\text{ALBERT}}$.
    \item \textbf{Transferability metrics} to new domains (e.g., heavy trucks, motorcycles).
\end{enumerate}



\end{document}